\documentclass[conference]{IEEEtran}
\IEEEoverridecommandlockouts
\usepackage{amsmath,amsfonts}
\usepackage[ruled,vlined,linesnumbered]{algorithm2e}
\usepackage{array}
\usepackage{subcaption}
\usepackage{textcomp}
\usepackage{stfloats}
\usepackage{url}
\usepackage{verbatim}
\usepackage{graphicx}
\usepackage{cite}
\usepackage{multirow}
\usepackage{lipsum}
\usepackage{mathtools}
\usepackage{cuted}
\usepackage{breqn}
\usepackage[english]{babel}
\usepackage[autostyle]{csquotes}
\usepackage[acronym,toc,shortcuts]{glossaries}
\usepackage{comment}
\usepackage{siunitx}
\usepackage{textgreek}
\usepackage{xcolor}
\usepackage{rotating}

\def\BibTeX{{\rm B\kern-.05em{\sc i\kern-.025em b}\kern-.08em
    T\kern-.1667em\lower.7ex\hbox{E}\kern-.125emX}}
\begin{document}

\title{Secure Hierarchical Federated Learning in Vehicular Networks Using Dynamic Client Selection and Anomaly Detection\\
\thanks{This work is supported by the Scientific and Technological Research Council of Turkey Grant number 119C058 and Ford Otosan.}
}

\author{\IEEEauthorblockN{ M. Saeid HaghighiFard}
\IEEEauthorblockA{\textit{Department of Electrical and Electronics Engineering} \\
\textit{Koç University}\\
Istanbul, Turkey \\
mhaghighifard21@ku.edu.tr}
\and
\IEEEauthorblockN{ Sinem Coleri}
\IEEEauthorblockA{\textit{Department of Electrical and Electronics Engineering} \\
\textit{Koç University}\\
Istanbul, Turkey \\
scoleri@ku.edu.tr}
}

\maketitle

\begin{abstract}
Hierarchical Federated Learning (HFL) faces the significant challenge of adversarial or unreliable vehicles in vehicular networks, which can compromise the model's integrity through misleading updates. Addressing this, our study introduces a novel framework that integrates dynamic vehicle selection and robust anomaly detection mechanisms, aiming to optimize participant selection and mitigate risks associated with malicious contributions. Our approach involves a comprehensive vehicle reliability assessment, considering historical accuracy, contribution frequency, and anomaly records. An anomaly detection algorithm is utilized to identify anomalous behavior by analyzing the cosine similarity of local or model parameters during the federated learning (FL) process. These anomaly records are then registered and combined with past performance for accuracy and contribution frequency to identify the most suitable vehicles for each learning round. Dynamic client selection and anomaly detection algorithms are deployed at different levels, including cluster heads (CHs), cluster members (CMs), and the Evolving Packet Core (EPC), to detect and filter out spurious updates. Through simulation-based performance evaluation, our proposed algorithm demonstrates remarkable resilience even under intense attack conditions. Even in the worst-case scenarios, it achieves convergence times at $63$\% as effective as those in scenarios without any attacks. Conversely, in scenarios without utilizing our proposed algorithm, there is a high likelihood of non-convergence in the FL process. 
\end{abstract}


\section{Introduction}
Vehicular Ad hoc Networks (VANETs) increasingly incorporate machine learning (ML) algorithms to enhance safety within transportation networks. These algorithms leverage data from sensors like LIDAR, RADAR, and vehicle cameras to facilitate informed decision-making, offering services such as location-based services, traffic flow prediction and control, and autonomous driving. Centralized ML approaches often involve training a neural network on extensive datasets from vehicle edge devices. Federated learning (FL), a newer approach, shifts ML to the edge to minimize transmission overhead and maintain user privacy. In FL, edge devices transmit parameter gradients to a central entity instead of local datasets. This entity aggregates these gradients to update the model parameters, which are then distributed back to the edge devices for further training, repeating the process until the model is fully trained~\cite{FLBook.ch2}.

Hierarchical Federated Learning (HFL) represents a recent advancement devised to address the challenges encountered in FL within vehicular networks. In this approach, a central cloud server orchestrates the training of a global model, leveraging edge servers as intermediaries for model aggregation. HFL emerges as a solution for overcoming challenges such as limited communication resources, vehicle mobility, and data heterogeneity inherent in vehicular networks~\cite{9148862,9054634}. HFL uses limited communication resources since the local parameters are transferred to the edge servers rather than relying solely on the cloud server and a limited number of edge servers connected to the cloud. Additionally, HFL handles mobility by expanding the coverage range for cloud servers through edge servers and managing data heterogeneity between edge servers, contributing to improved learning performance. For implementation of the HFL over vehicular networks,~\cite{chen2024mobility} leverages mobility to enhance learning performance by alleviating data heterogeneity.~\cite{chen2021semiasynchronous} proposes semi-asynchronous HFL for Cooperative Intelligent Transportation Systems (C-ITS) to optimize communication efficiency, particularly in the presence of heterogeneous road vehicles. In addition,~\cite{haghighifard2024hierarchical} proposes an HFL approach over multi-hop clustered VANET. This method incorporates a practical clustering algorithm and metrics tailored to handle vehicle mobility and non-IID data structures. However, none of these studies on HFL in vehicular networks address the security attack problem. A model poisoning attack occurs when an attacker directly manipulates the local model parameters sent from the vehicle to the CH or EPC, often by adding random noise to the local model. This direct intervention can significantly hinder the global model's ability to converge effectively~\cite{10024252,10274102,9857216}.

Anomaly detection mechanisms have been developed against model poisoning attacks in FL to identify distinguishable differences between malicious and benign models~\cite {9751538,sun2021flwbc,9734158}.~\cite{9751538} introduces a Differential Privacy-exploited stealthy model poisoning (DeSMP) attack targeting FL models and proposes a defense strategy based on reinforcement learning (RL) to adjust the privacy level of FL models.~\cite{sun2021flwbc} proposes a client-based defense, known as White Blood Cell for Federated Learning (FL-WBC), which aims to identify the parameter space where long-lasting attack effects on parameters reside and perturb that space during local training.~\cite{9734158} enables the filtering out of anomalous models by assessing the similarity in historical changes across clients. None of the existing methods designed to counteract malicious users and attacks in FL have been applied to HFL within dynamic vehicular environments and real-world scenarios. 

Dynamic client selection follows anomaly detection in FL to optimize the selection of participating vehicles for each learning round based on the output of the anomaly detection~\cite{Khorramfar_2023, yaldiz2023secure}.~\cite{Khorramfar_2023} introduces the Credit-Based Client Selection (CBCS). CBCS, utilizing a credit system, assesses clients on the accuracy of their historical model contributions, favoring those with higher scores for model updates and scrutinizing or excluding lower-scored clients.~\cite{ yaldiz2023secure} presents CosDefense, a detection algorithm based on cosine similarity. CosDefense utilizes the cosine similarity of the last layer's weight between the global model and each local update as a critical indicator to identify malicious model updates. The server computes the cosine similarity score for each client update against the global model, isolating and excluding any client with a significantly higher score, indicating potential malicious intent. None of these studies, however, explore the behavior of dynamic client selection within the dynamic context of vehicular networks and overlook the HFL framework.

This paper introduces an innovative framework that combines dynamic client selection with anomaly detection in HFL over cluster-based VANET, specifically designed to enhance the model's accuracy and reliability in an unsecured environment. The original contributions of the paper are listed as follows:
\begin{itemize}
\item We propose a novel framework to provide robust protection against security vulnerabilities for executing HFL within vehicular networks, for the first time in the literature. Within this framework, vehicles form clusters to implement HFL, where a reliability score is used to identify and prioritize well-behaved vehicles for participation in the learning process. Conversely, vehicles exhibiting malicious or suspicious behavior are temporarily excluded, with their behavior reassessed after a predefined period. This strategy ensures the integrity of the learning process while promoting a high-quality, secure, collaborative learning environment.
\item We propose a novel dynamic client selection methodology integrating historical accuracy, contribution frequency, and anomaly records as key metrics for executing HFL within vehicular networks, for the first time in the literature. Historical accuracy facilitates prioritizing vehicles with a demonstrated ability to cooperate reliably in the FL process. Contribution frequency assists in identifying vehicles with superior communication links and hardware, thus reducing packet loss. Anomaly records aid in identifying vehicles exhibiting minimal or no anomalous behavior. Leveraging these metrics allows us to select the most accurate and reliable vehicles for the FL process.
\item We demonstrate via extensive simulations that our proposed methodology significantly outperforms previously proposed no clustering-based and alternative attacker detection algorithms in terms of both convergence time and resilience across various attack scenarios.
\end{itemize}
The structure of this paper is organized as follows: Section II describes the system model. Section III gives the proposed secure HFL framework, incorporating dynamic client selection and anomaly detection algorithms. Section IV provides the performance evaluation of the proposed algorithm. The paper concludes with Section V, which summarizes the findings and outlines potential future research.

\section{System Model}

The network consists of cluster heads (CHs), cluster members (CMs), and the Evolving Packet Core (EPC). Vehicles are equipped with two communication interfaces for vehicle-to-vehicle (V2V) and vehicle-to-infrastructure or vehicle-to-network (V2I/V2N) communications, using standards such as IEEE 802.11p \cite{5888501}, IEEE 802.11bd for Wireless Access in Vehicular Environments (WAVE)~\cite{9779322}, LTE D2D~\cite{7497762}, and 5G NR V2X~\cite{9392787}.

Vehicles are organized into clusters dynamically based on the Cluster-based Hierarchical Federated Learning (CbHFL)~\cite{haghighifard2024hierarchical} approach, with roles as CHs and CMs changing over time. The nodes are clustered by using metrics combining average relative speed and similarity of local model parameters. A HFL algorithm is implemented over this clustered network. CMs act as clients, providing model parameter updates to their CHs based on local data gathered in a non-IID distribution structure. CHs, acting as edge nodes, aggregate these updates and send them periodically to the EPC for general model aggregation. The EPC then disseminates the latest updates back to the CHs, which then transmit them to the CMs. These updates are calculated using stochastic gradient descent (SGD).

Each vehicle has a Vehicle Information Base (VIB), a repository storing crucial vehicle parameters including direction, location, velocity, clustering state of the vehicle, model parameters, number of communication rounds, all possible contributions, the accuracy of contributions, anomaly records, and the reliability score. The VIB is updated whenever a vehicle's information changes or receives a "$HELLO\_PACKET$" or model parameter request. A "$HELLO\_PACKET$" contains essential information, including 
direction, location, velocity, current clustering state, ID of the vehicle that connects the node to the cluster, cosine similarity of FL model parameters, and average relative speed with respect to all neighboring nodes.
Each vehicle's VIB continuously monitors and updates these elements. Each node in the network must have sufficient storage and computing power to store and process FL model parameters.

Certain vehicles in the network face difficulties reliably transmitting packets to other vehicles or the EPC, a problem attributed to the characteristics of the communication channel. Additionally, a group of vehicles intentionally introduces Gaussian noise into the model parameters. This behavior indicates malicious intent, as these vehicles actively seek to disrupt the integrity of the system's data and operations.

CHs choose vehicles within the network based on a predefined percentage, employing a reliability score as the selection criterion. This method ensures that vehicle selection for network tasks is both data-driven and reflective of their past performance and reliability.

\section{Secure HFL using Dynamic Client Selection and Anomaly Detection}

We introduce a novel approach for dynamic client selection coupled with an anomaly detection algorithm tailored for CbHFL. For anomaly detection, we employ cosine similarity to detect any irregularities in the local or model parameters during the FL process. The evaluation of a vehicle's reliability involves combining its anomaly history with historical accuracy and contribution frequency. Vehicles exhibiting high reliability are prioritized in contributing to the FL process, ensuring its effectiveness through the expertise and dependability of these vehicles.

Each vehicle in the network is allocated a comprehensive reliability score, determined by combining historical accuracy, contribution frequency, and anomaly record metrics, as specified below: 

\begin{itemize}
\item \textbf{Historical Accuracy}: This metric evaluates the average performance of a vehicle's accuracy over time. The EPC and all CHs calculate this metric by analyzing the accuracy of a vehicle's model across multiple communication rounds on a validation set. The validation set is used to assess model performance while adjusting hyper-parameters. Accuracy is typically assessed using evaluation metrics to measure the model's performance on the validation dataset. This metric quantifies the percentage of accurate predictions made by the learning model when tested with samples from the dataset~\cite{Goodfellow2016}. High historical accuracy suggests that the vehicle consistently contributes beneficial updates to the global model. The historical accuracy of vehicle $k$ is calculated by
\begin{multline*} \label{eq:1}
    Historical\_Accuracy_{{k}} = Total\_Accuracy_{k,i}/i = \\
    (1/i) * \sum_{n=1}^{i}(Accuracy\_of\_Contribution_{k,n}),\tag{1}
\end{multline*}
where $i$ represents the $i$th communication round considered as a single FL iteration within the FL procedure, $Accuracy\_of\_Contribution_{k,n}$ is the accuracy level of $k$th vehicle as determined by the CH or EPC calculation for the $n$th round, and
$Total\_Accuracy_{k,i}$ represents the cumulative sum of all $Accuracy\_of\_Contribution_{k,n}$ values across all communication rounds up to the current round $i$.
\item \textbf{Contribution Frequency}: This aspect considers how often a vehicle participates in the FL rounds. A higher frequency of contributions can indicate a more engaged and reliable participant. However, this metric balances the quality of contributions together with historical accuracy, ensuring that frequent participation does not overshadow the importance of valuable data. The contribution frequency of vehicle $k$ is calculated by
\begin{multline*} \label{eq:2}
    Contribution\_Freq_{k} = \\
    {Total\_Contributions_{k}}/{i},\tag{2}
\end{multline*}
where $Total\_Contributions_{k}$ represents the number of contributions made by vehicle $k$ across all communication rounds up to the current round $i$ in the FL process.
\item \textbf{Anomaly Record}: Anomaly detection plays a crucial role in identifying and mitigating the impact of adversarial attacks. By keeping track of the number of times a vehicle’s contributions have been flagged as anomalies through cosine similarity measurements between successive local parameters, we can assess its risk profile. A high anomaly record may indicate a compromised or unreliable vehicle, necessitating a lower reliability score to safeguard the learning process. The Anomaly record of vehicle $k$ is calculated by
\begin{multline*} \label{eq:3}
    Anomaly\_Record_{k} = \\
    {Total\_Anomalous\_Contributions_{k}}/{i},\tag{3}
\end{multline*}

where $Total\_Anomalous\_Contributions_{k}$ refers to the total number of the instances where $k$th vehicle contributions have been flagged as anomalous across all communication rounds up to the current round $i$.

\end{itemize}

The final reliability score for each vehicle is determined by a weighted sum of the three metrics as given below:
\begin{multline*} \label{eq:4}
Reliability\_Score_{k} = \\
({Accuracy\_Weight \times Historical\_Accuracy_{k}})+\\
({Frequency\_Weight \times Contribution\_Freq_{k}})-\\
({Anomaly\_Weight \times Anomaly\_Record_{k}})\tag{4}
\end{multline*}

Assigning weights to different metrics for calculating vehicle reliability allows a flexible and context-sensitive evaluation of each vehicle's contribution in FL. This weighting system can be tailored to prioritize certain aspects of reliability based
on the specific needs and goals of the FL application. This level of customization is crucial for optimizing the learning process in diverse and dynamic environments like vehicular networks.

The reliability score is dynamically updated as the network evolves, ensuring that the selection process remains adaptive to changes in vehicle behavior and network conditions. Vehicles with higher reliability scores are prioritized for participation in the model training rounds, enhancing the overall robustness and accuracy of the global model.

\begin{algorithm*}[!ht]
Initialize $SELECTED\_CLIENT\_PERCENTAGE$ and $Unblock\_Time$;\\
\For {all $CM \in VIB$}{
    Initialize $Reliability\_Score_{CM}$ based on VIB;\\
}
    Sort all $CM \in \mathcal{CM}$ by $Reliability\_Score$s in descending order;\\
    Select top $SELECTED\_CLIENT\_PERCENTAGE$ of vehicles in Cluster as $Selected\_Clients$;\\
\For {all $CM \in Selected\_Clients$}
    {
        \If {$Block\_Flag_{CM} =$ $TRUE$ $\wedge$ $block\_duration_{CM}$ $< Unblock\_Time$}{
          $block\_duration_{CM}$ ++;\\
            Skip current vehicle;}
            \Else{
            Reset $Block\_Flag_{CM}$ to $FALSE$;}

    \If {$Cosine\_Similarity(g_{{CM},i},g_{{CM}, i-1})$ $<$ $0.5$}{
        Set $Block\_Flag_{CM} =$ $TRUE$ and $block\_duration_{CM}=0$;\\
        $Total\_Anomalous\_Contributions_{CM}$ ++;
    }
    \Else{
        $Total\_Contributions_{CM}$ ++;\\
        Calculate $Accuracy\_of\_Contribution_{{CM},i}$;\\
        $Total\_Accuracy_{{CM},i} = Accuracy\_of\_Contribution_{{CM},i} + Total\_Accuracy_{{CM},i-1}$;\\
        Calculate $Historical\_Accuracy_{CM} = Total\_Accuracy_{{CM},i} / i$;\\
        Calculate $Contribution\_Freq_{CM} = Total\_Contributions_{CM} / i$;\\
        Calculate $Anomaly\_Record_{CM} = Total\_Anomalous\_Contributions_{CM} / i$;\\
        $Reliability\_Score_{CM}$ = $(Accuracy\_Weight* Historical\_Accuracy_{CM})$
        $ + (Frequency\_Weight * Contribution\_Freq_{CM})$
        $- (Anomaly\_Weight * Anomaly\_Record_{CM})$;\\
        Add $Block\_Flag_{CM}$, $block\_duration_{CM}$, $Total\_Accuracy_{{CM},i}$, $Total\_contributions_{CM}$, $Total\_Anomalous\_Contributions_{CM}$, and $Reliability\_Score_{CM}$ to the VIB;\\
    }
    }

\caption{Dynamic Client Selection and Anomaly Detection at the CH}
\end{algorithm*}

Dynamic client selection and anomaly detection at the CH level are detailed in Algorithm 1. This algorithm outlines the process for identifying and selecting vehicles based on dynamically assessed criteria, ensuring effective and reliable participation in the network. The algorithm's core lies in selecting vehicles for each round of FL. CHs are assigned the roles of identifying anomalies and computing vehicle reliability scores of the cluster members, referred to as $CM$, within their respective cluster $\mathcal{CM}$. The CH first initializes the $Reliability\_Score_{CM}$ for the cluster members based on VIB. Subsequently, it sorts the vehicles by their $Reliability\_Score$s in descending order. Then, based on a predefined $SELECTED\_CLIENT\_PERCENTAGE$, the algorithm chooses the top segment of these vehicles for participation in the upcoming training round, ensuring only the most reliable vehicles are selected (Lines $1$-$5$). The CH then evaluates whether each selected vehicle should be allowed to participate in the training. It checks the $Block\_Flag_{CM}$ and the $block\_duration_{CM}$ of every CM based on the information received from previous communication rounds or previous CH or EPC. The $Block\_Flag_{CM}$ serves as a binary indicator to identify CM exhibiting anomalous behavior, while $block\_duration_{CM}$ represents the time duration during which the blocking remains. Once anomalous behavior is detected, the CH waits for the $Unblock\_Time$ period to elapse before reassessing the CM as part of the cluster, considering its $Reliability\_Score_{CM}$. The $Unblock\_Time$ refers to a preset duration designated for penalizing vehicles exhibiting undesirable behavior. If a vehicle is flagged and its block duration does not exceed $Unblock\_Time$, the algorithm skips this vehicle and proceeds to the next (Lines $6$-$11$).
The algorithm assesses the consistency of the vehicle's contributions by comparing the similarity between a vehicle's current and previous gradient updates, denoted as $g_{CM,i}$ and $g_{CM,i-1}$, respectively. These updates are the outputs of SGD at each CM, transmitted to its corresponding CH. Including cosine similarity as a criterion introduces a robust anomaly detection mechanism. A threshold of 0.5 is used to determine benign versus anomalous behavior. If a vehicle's updates are found to be inconsistent (i.e., the cosine similarity is below 0.5), it is marked by setting the $Block\_Flag_{CM}$ to $TRUE$, and its $Total\_Anomalous\_Contributions_{CM}$ count is increased by $1$ (Lines $12$-$14$).
The $Reliability\_Score_{CM}$ of each vehicle is recalculated in every round. It is a weighted sum of  $Historical\_Accuracy_{CM}$, $Contribution\_Freq_{CM}$, and $Anomaly\_Record_{CM}$, each multiplied by its respective weight. All calculated factors are incorporated into VIB to facilitate future calculations (Lines $15$-$24$).

\begin{algorithm*}[!ht]
\For {all $CH \in VIB$}
{
    \If {$Block\_Flag_{CH} =$ $TRUE$ $\wedge$ $block\_duration_{CH}$ $< Unblock\_Time$}{
            $block\_duration_{CH}$ ++;\\
            Skip current vehicle;}
     \Else{
            Reset $Block\_Flag_{CH}$ to $FALSE$;}
    \If {$Cosine\_Similarity(\theta_{CH,i},\theta_{CH,i-1})$ $<$ $0.5$}{
        Set $Block\_Flag_{CH} =$ $TRUE$  and $block\_duration_{CH}=0$;\\
        $Total\_Anomalous\_Contributions_{CH}$++;\\
    }
    \Else{
        $Total\_Contributions_{CH}$ ++;\\
        Calculate $Accuracy\_of\_Contribution_{{CH},i}$;\\
        $Total\_Accuracy_{{CH},i} = Accuracy\_of\_Contribution_{{CH},i} + Total\_Accuracy_{{CH},i-1}$;\\
        Calculate $Historical\_Accuracy_{CH} = Total\_Accuracy_{{CH},i} / i$;\\
        Calculate $Contribution\_Freq_{CH} = Total\_Contributions_{CH} / i$;\\
        Calculate $Anomaly\_Record_{CH} = Total\_Anomalous\_Contributions_{CH} / i$;\\
        $Reliability\_Score_{CH}$ = $(Accuracy\_Weight* Historical\_Accuracy_{CH})$
        $ + (Frequency\_Weight * Contribution\_Freq_{CH})$
        $- (Anomaly\_Weight * Anomaly\_Record_{CH})$;\\
        Add $Block\_Flag_{CH}$, $block\_duration_{CH}$, $Total\_Accuracy_{{CH},i}$, $Total\_contributions_{CH}$, $Total\_Anomalous\_Contributions_{CH}$, and $Reliability\_Score_{CH}$ to the VIB;\\
    }
}
\caption{Anomaly Detection for CHs at EPC}
\end{algorithm*}

Algorithm 2 focuses on anomaly detection at the EPC for CHs. It aims to identify anomalies and compute reliability scores for CHs using the VIB. Assessing a CH's suitability for collaboration entails evaluating multiple factors by the EPC. These factors include analyzing the $Block\_Flag_{CH}$ and $block\_duration_{CH}$ linked to each CH interaction. Upon detecting anomalous behavior, the EPC temporarily bypasses that CH until the $Unblock\_Time$ period elapses. Subsequently, the EPC reassesses the CH's suitability for participation. If a vehicle remains flagged and its block duration is less than $Unblock\_Time$, the algorithm proceeds to the next vehicle. Conversely, if no anomalies are detected, the flag is reset (Lines $1$-$6$). For each CH, the algorithm assesses the cosine similarity between its current model parameters ($\theta_{CH,i}$) and the ones from the previous iteration ($\theta_{CH,i-1}$). A cosine similarity below 0.5, indicating a notable alteration or inconsistency in model parameters, triggers the setting of the $Block\_Flag_{CH}$ to $TRUE$ for the implicated CH, increasing its $Total\_Anomalous\_Contributions_{CH}$ by $1$ (Lines $7$-$9$). The $Reliability\_Score_{CH}$ for each vehicle is recalculated each round, based on a weighted combination of $Historical\_Accuracy_{CH}$, $Contribution\_Freq_{CH}$, and $Anomaly\_Record_{CH}$, with each factor assigned a specific weight. These calculated elements are then added to the VIB for subsequent analysis and decision-making (Lines $10$-$18$).

\section{Performance Evaluation}

\begin{table*}[ht]
\centering
\caption{Convergence time based on different values of mean, variance, and $\epsilon$ for attack only on the 10$th$ round}
\begin{tabular}{|cc|cccc|}
\hline
\multicolumn{2}{|c|}{\multirow{2}{*}{}}                                                                                 & \multicolumn{4}{c|}{Attack only on 10$th$ Round}                                                                                                                                                                                                                                                 \\ \cline{3-6} 
\multicolumn{2}{|c|}{}                                                                                                  & \multicolumn{1}{c|}{CosDefense} & \multicolumn{1}{c|}{\begin{tabular}[c]{@{}c@{}}No Clustering+\\ Proposed Algorithm\end{tabular}} & \multicolumn{1}{c|}{\begin{tabular}[c]{@{}c@{}}CbHFL+\\ Proposed Algorithm\end{tabular}} & \begin{tabular}[c]{@{}c@{}}CbHFL\\ with No Attack\end{tabular} \\ \hline
\multicolumn{1}{|c|}{\multirow{2}{*}{\begin{tabular}[c]{@{}c@{}}Mean= 0\\      Var = 0.1\end{tabular}}} & $\epsilon= 0.1$  & \multicolumn{1}{c|}{26}         & \multicolumn{1}{c|}{27}                                                                          & \multicolumn{1}{c|}{15}                                                                  & 13                                                             \\ \cline{2-6} 
\multicolumn{1}{|c|}{}                                                                                  & $\epsilon= 0.01$ & \multicolumn{1}{c|}{49}         & \multicolumn{1}{c|}{47}                                                                          & \multicolumn{1}{c|}{26}                                                                  & 26                                                             \\ \hline
\multicolumn{1}{|c|}{\multirow{2}{*}{\begin{tabular}[c]{@{}c@{}}Mean= 0\\      Var = 0.2\end{tabular}}} & $\epsilon= 0.1$  & \multicolumn{1}{c|}{15}         & \multicolumn{1}{c|}{15}                                                                          & \multicolumn{1}{c|}{13}                                                                  & 10                                                             \\ \cline{2-6} 
\multicolumn{1}{|c|}{}                                                                                  & $\epsilon= 0.01$ & \multicolumn{1}{c|}{$\infty$}        & \multicolumn{1}{c|}{$\infty$}                                                                         & \multicolumn{1}{c|}{64}                                                                  & 56                                                             \\ \hline
\multicolumn{1}{|c|}{\multirow{2}{*}{\begin{tabular}[c]{@{}c@{}}Mean= 0\\      Var = 0.3\end{tabular}}} & $\epsilon= 0.1$  & \multicolumn{1}{c|}{32}         & \multicolumn{1}{c|}{32}                                                                          & \multicolumn{1}{c|}{23}                                                                  & 20                                                             \\ \cline{2-6} 
\multicolumn{1}{|c|}{}                                                                                  & $\epsilon= 0.01$  & \multicolumn{1}{c|}{45}         & \multicolumn{1}{c|}{41}                                                                          & \multicolumn{1}{c|}{28}                                                                  & 24                                                             \\ \hline
\multicolumn{1}{|c|}{\multirow{2}{*}{\begin{tabular}[c]{@{}c@{}}Mean= 1\\      Var = 0.1\end{tabular}}} & $\epsilon= 0.1$  & \multicolumn{1}{c|}{28}         & \multicolumn{1}{c|}{28}                                                                          & \multicolumn{1}{c|}{11}                                                                  & 10                                                             \\ \cline{2-6} 
\multicolumn{1}{|c|}{}                                                                                  & $\epsilon= 0.01$ & \multicolumn{1}{c|}{50}         & \multicolumn{1}{c|}{47}                                                                          & \multicolumn{1}{c|}{33}                                                                  & 28                                                             \\ \hline
\multicolumn{1}{|c|}{\multirow{2}{*}{\begin{tabular}[c]{@{}c@{}}Mean= 1\\      Var = 0.2\end{tabular}}} & $\epsilon= 0.1$  & \multicolumn{1}{c|}{$\infty$}        & \multicolumn{1}{c|}{$\infty$}                                                                         & \multicolumn{1}{c|}{31}                                                                  & 25                                                             \\ \cline{2-6} 
\multicolumn{1}{|c|}{}                                                                                  & $\epsilon= 0.01$ & \multicolumn{1}{c|}{$\infty$}        & \multicolumn{1}{c|}{$\infty$}                                                                         & \multicolumn{1}{c|}{31}                                                                  & 31                                                             \\ \hline
\multicolumn{1}{|c|}{\multirow{2}{*}{\begin{tabular}[c]{@{}c@{}}Mean= 1\\      Var = 0.3\end{tabular}}} & $\epsilon= 0.1$  & \multicolumn{1}{c|}{45}         & \multicolumn{1}{c|}{40}                                                                          & \multicolumn{1}{c|}{25}                                                                  & 20                                                             \\ \cline{2-6} 
\multicolumn{1}{|c|}{}                                                                                  & $\epsilon= 0.01$ & \multicolumn{1}{c|}{$\infty$}        & \multicolumn{1}{c|}{$\infty$}                                                                         & \multicolumn{1}{c|}{39}                                                                  & 39                                                             \\ \hline
\multicolumn{1}{|c|}{\multirow{2}{*}{\begin{tabular}[c]{@{}c@{}}Mean= 2\\      Var = 0.1\end{tabular}}} & $\epsilon= 0.1$  & \multicolumn{1}{c|}{35}         & \multicolumn{1}{c|}{38}                                                                          & \multicolumn{1}{c|}{29}                                                                  & 18                                                             \\ \cline{2-6} 
\multicolumn{1}{|c|}{}                                                                                  & $\epsilon= 0.01$ & \multicolumn{1}{c|}{$\infty$}        & \multicolumn{1}{c|}{$\infty$}                                                                         & \multicolumn{1}{c|}{61}                                                                  & 61                                                             \\ \hline
\multicolumn{1}{|c|}{\multirow{2}{*}{\begin{tabular}[c]{@{}c@{}}Mean= 2\\      Var = 0.2\end{tabular}}} & $\epsilon= 0.1$  & \multicolumn{1}{c|}{$\infty$}        & \multicolumn{1}{c|}{$\infty$}                                                                         & \multicolumn{1}{c|}{35}                                                                  & 11                                                             \\ \cline{2-6} 
\multicolumn{1}{|c|}{}                                                                                  & $\epsilon= 0.01$ & \multicolumn{1}{c|}{$\infty$}        & \multicolumn{1}{c|}{$\infty$}                                                                         & \multicolumn{1}{c|}{62}                                                                  & 25                                                             \\ \hline
\multicolumn{1}{|c|}{\multirow{2}{*}{\begin{tabular}[c]{@{}c@{}}Mean= 2\\      Var = 0.3\end{tabular}}} & $\epsilon= 0.1$  & \multicolumn{1}{c|}{$\infty$}        & \multicolumn{1}{c|}{60}                                                                          & \multicolumn{1}{c|}{25}                                                                  & 21                                                             \\ \cline{2-6} 
\multicolumn{1}{|c|}{}                                                                                  & $\epsilon= 0.01$ & \multicolumn{1}{c|}{$\infty$}        & \multicolumn{1}{c|}{$\infty$}                                                                         & \multicolumn{1}{c|}{45}                                                                  & 36                                                             \\ \hline
\end{tabular}
\end{table*}

\begin{table*}[]
\centering
\caption{Convergence time based on different values of mean, variance, and $\epsilon$ for continuous attack}
\begin{tabular}{|cc|cccc|}
\hline
\multicolumn{2}{|c|}{\multirow{2}{*}{}}                                                                                 & \multicolumn{4}{c|}{Continuous Attack}                                                                                                                                                                                                                                                         \\ \cline{3-6} 
\multicolumn{2}{|c|}{}                                                                                                  & \multicolumn{1}{c|}{CosDefense} & \multicolumn{1}{c|}{\begin{tabular}[c]{@{}c@{}}No Clustering+\\ Proposed Algorithm\end{tabular}} & \multicolumn{1}{c|}{\begin{tabular}[c]{@{}c@{}}CbHFL+\\ Proposed Algorithm\end{tabular}} & \begin{tabular}[c]{@{}c@{}}CbHFL\\ with No Attack\end{tabular} \\ \hline
\multicolumn{1}{|c|}{\multirow{2}{*}{\begin{tabular}[c]{@{}c@{}}Mean= 0\\      Var = 0.1\end{tabular}}} & $\epsilon= 0.1$  & \multicolumn{1}{c|}{54}         & \multicolumn{1}{c|}{48}                                                                          & \multicolumn{1}{c|}{26}                                                                  & 15                                                             \\ \cline{2-6} 
\multicolumn{1}{|c|}{}                                                                                  & $\epsilon= 0.01$ & \multicolumn{1}{c|}{58}         & \multicolumn{1}{c|}{58}                                                                          & \multicolumn{1}{c|}{35}                                                                  & 35                                                             \\ \hline
\multicolumn{1}{|c|}{\multirow{2}{*}{\begin{tabular}[c]{@{}c@{}}Mean= 0\\      Var = 0.2\end{tabular}}} & $\epsilon= 0.1$  & \multicolumn{1}{c|}{42}         & \multicolumn{1}{c|}{45}                                                                          & \multicolumn{1}{c|}{30}                                                                  & 21                                                             \\ \cline{2-6} 
\multicolumn{1}{|c|}{}                                                                                  & $\epsilon= 0.01$ & \multicolumn{1}{c|}{$\infty$}        & \multicolumn{1}{c|}{45}                                                                          & \multicolumn{1}{c|}{36}                                                                  & 25                                                             \\ \hline
\multicolumn{1}{|c|}{\multirow{2}{*}{\begin{tabular}[c]{@{}c@{}}Mean= 0\\      Var = 0.3\end{tabular}}} & $\epsilon= 0.1$  & \multicolumn{1}{c|}{48}         & \multicolumn{1}{c|}{48}                                                                          & \multicolumn{1}{c|}{33}                                                                  & 14                                                             \\ \cline{2-6} 
\multicolumn{1}{|c|}{}                                                                                  & $\epsilon= 0.01$  & \multicolumn{1}{c|}{$\infty$}        & \multicolumn{1}{c|}{$\infty$}                                                                         & \multicolumn{1}{c|}{46}                                                                  & 36                                                             \\ \hline
\multicolumn{1}{|c|}{\multirow{2}{*}{\begin{tabular}[c]{@{}c@{}}Mean= 1\\      Var = 0.1\end{tabular}}} & $\epsilon= 0.1$  & \multicolumn{1}{c|}{52}         & \multicolumn{1}{c|}{60}                                                                          & \multicolumn{1}{c|}{32}                                                                  & 15                                                             \\ \cline{2-6} 
\multicolumn{1}{|c|}{}                                                                                  & $\epsilon= 0.01$ & \multicolumn{1}{c|}{64}         & \multicolumn{1}{c|}{60}                                                                          & \multicolumn{1}{c|}{32}                                                                  & 26                                                             \\ \hline
\multicolumn{1}{|c|}{\multirow{2}{*}{\begin{tabular}[c]{@{}c@{}}Mean= 1\\      Var = 0.2\end{tabular}}} & $\epsilon= 0.1$  & \multicolumn{1}{c|}{51}         & \multicolumn{1}{c|}{48}                                                                          & \multicolumn{1}{c|}{21}                                                                  & 15                                                             \\ \cline{2-6} 
\multicolumn{1}{|c|}{}                                                                                  & $\epsilon= 0.01$ & \multicolumn{1}{c|}{$\infty$}        & \multicolumn{1}{c|}{48}                                                                          & \multicolumn{1}{c|}{23}                                                                  & 26                                                             \\ \hline
\multicolumn{1}{|c|}{\multirow{2}{*}{\begin{tabular}[c]{@{}c@{}}Mean= 1\\      Var = 0.3\end{tabular}}} & $\epsilon= 0.1$  & \multicolumn{1}{c|}{64}         & \multicolumn{1}{c|}{38}                                                                          & \multicolumn{1}{c|}{26}                                                                  & 16                                                             \\ \cline{2-6} 
\multicolumn{1}{|c|}{}                                                                                  & $\epsilon= 0.01$ & \multicolumn{1}{c|}{$\infty$}        & \multicolumn{1}{c|}{$\infty$}                                                                         & \multicolumn{1}{c|}{58}                                                                  & 32                                                             \\ \hline
\multicolumn{1}{|c|}{\multirow{2}{*}{\begin{tabular}[c]{@{}c@{}}Mean= 2\\      Var = 0.1\end{tabular}}} & $\epsilon= 0.1$  & \multicolumn{1}{c|}{61}         & \multicolumn{1}{c|}{59}                                                                          & \multicolumn{1}{c|}{26}                                                                  & 15                                                             \\ \cline{2-6} 
\multicolumn{1}{|c|}{}                                                                                  & $\epsilon= 0.01$ & \multicolumn{1}{c|}{$\infty$}        & \multicolumn{1}{c|}{$\infty$}                                                                         & \multicolumn{1}{c|}{32}                                                                  & 19                                                             \\ \hline
\multicolumn{1}{|c|}{\multirow{2}{*}{\begin{tabular}[c]{@{}c@{}}Mean= 2\\      Var = 0.2\end{tabular}}} & $\epsilon= 0.1$  & \multicolumn{1}{c|}{65}         & \multicolumn{1}{c|}{57}                                                                          & \multicolumn{1}{c|}{50}                                                                  & 23                                                             \\ \cline{2-6} 
\multicolumn{1}{|c|}{}                                                                                  & $\epsilon= 0.01$ & \multicolumn{1}{c|}{$\infty$}        & \multicolumn{1}{c|}{$\infty$}                                                                         & \multicolumn{1}{c|}{59}                                                                  & 29                                                             \\ \hline
\multicolumn{1}{|c|}{\multirow{2}{*}{\begin{tabular}[c]{@{}c@{}}Mean= 2\\      Var = 0.3\end{tabular}}} & $\epsilon= 0.1$  & \multicolumn{1}{c|}{$\infty$}        & \multicolumn{1}{c|}{$\infty$}                                                                         & \multicolumn{1}{c|}{64}                                                                  & 25                                                             \\ \cline{2-6} 
\multicolumn{1}{|c|}{}                                                                                  & $\epsilon= 0.01$ & \multicolumn{1}{c|}{$\infty$}        & \multicolumn{1}{c|}{$\infty$}                                                                         & \multicolumn{1}{c|}{64}                                                                  & 53                                                             \\ \hline
\end{tabular}
\end{table*}

The simulations aim to evaluate the effectiveness of the proposed security algorithm when integrated with CbHFL, as compared to three approaches: First, a no clustering-based algorithm~\cite{9360666} combined with our proposed security algorithm; second, a cosine similarity-based attacker detection algorithm (CosDefense)~\cite{yaldiz2023secure}; and third, a scenario with no attacks. In the no clustering-based algorithm, clustering is not employed, treating all vehicles as clients with the EPC serving as the central server for data processing. By combining our proposed algorithm with the no clustering approach, we aim to assess the clustering and HFL performance. The CosDefense algorithm, which also operates without clustering, distinguishes itself by computing the cosine similarity score of the last layer's weight between the global model and each client update. It identifies potentially malicious vehicles based on scores significantly higher than the average and excludes them from model aggregation in each round. CosDefense has not explored integrating clustering algorithms with anomaly detection. Our proposed algorithm incorporates both clustering and cosine similarity metrics for enhanced anomaly detection. Finally, by comparing all of these algorithms against a scenario with no attacks, we aim to demonstrate how CbHFL can be improved. This comparison seeks to establish which approach is most effective at maintaining performance in the face of attacks.

\subsection{Simulation Setup}
The simulations utilize Python, with vehicle mobility generated by Simulation of Urban Mobility (SUMO) and data streaming via KAFKA. This setup creates a realistic vehicular network scenario for FL. SUMO models individual driver behaviors and traffic dynamics, while KAFKA is utilized to transmit packets and model parameters, enabling communication among vehicles. FL models are built using PyTorch. 

The simulation environment replicates one km² area with two-lane roads, where vehicle movements are simulated using a Poisson process, incorporating a variety of vehicles with speeds ranging from 10 m/s to 35 m/s. For communication, IEEE802.11p and 5G NR are employed for vehicle-to-vehicle and vehicle-to-infrastructure interactions, respectively. The Winner+ B1 propagation model, as proposed in~\cite{9599363}, is utilized for vehicle-to-vehicle communications. Meanwhile, the Friis propagation model is applied in vehicle-to-infrastructure communication scenarios to estimate signal attenuation and propagation characteristics between vehicles and 5G NR base stations, as detailed in~\cite{3gpptr38901}. The FL training leverages the MNIST dataset to train SGD, focusing on non-IID data distribution among vehicles. The main evaluation metrics are accuracy, indicating the model's predictive performance\cite{Goodfellow2016}, and convergence time, measuring the speed of reaching desired learning outcomes~\cite{mcmahan2017communication}. This setup aims to explore the efficiency of FL in vehicular networks under realistic conditions.

In the consistently applied scenario, there are 25 vehicles, each with a transmission range of 100 meters. Among these, 20\% are designated as attackers, employing fake additive Gaussian noise with varying means and variances across two distinct models: one model experiences an attack that only happens on the 10th round. In contrast, in the other model, attacks are continuously carried out by malicious vehicles from the 10th round. The rationale behind initiating the attack from this particular round is based on the assumption that all vehicles have entered the system and are expected to be stable. The maximum number of hops allowed is set to one, indicating that communication can only occur directly between CM and CH without intermediary hops. The proportion of selected clients for each training round is set to 75\%, with an unblock time fixed at 5. The weighting of each parameter towards the reliability score is equal, implying that all parameters contribute identically to the overall reliability evaluation. This setup is designed to assess the robustness of FL models against adversarial tactics, explicitly observing the impact of sporadic and continuous network attacks on model integrity and performance.

\subsection{Performance Evaluation of Proposed Algorithm}

\begin{figure}[!t]
     \centering
     \begin{subfigure}{0.45\textwidth}
         \centering
         \includegraphics[width=\textwidth]{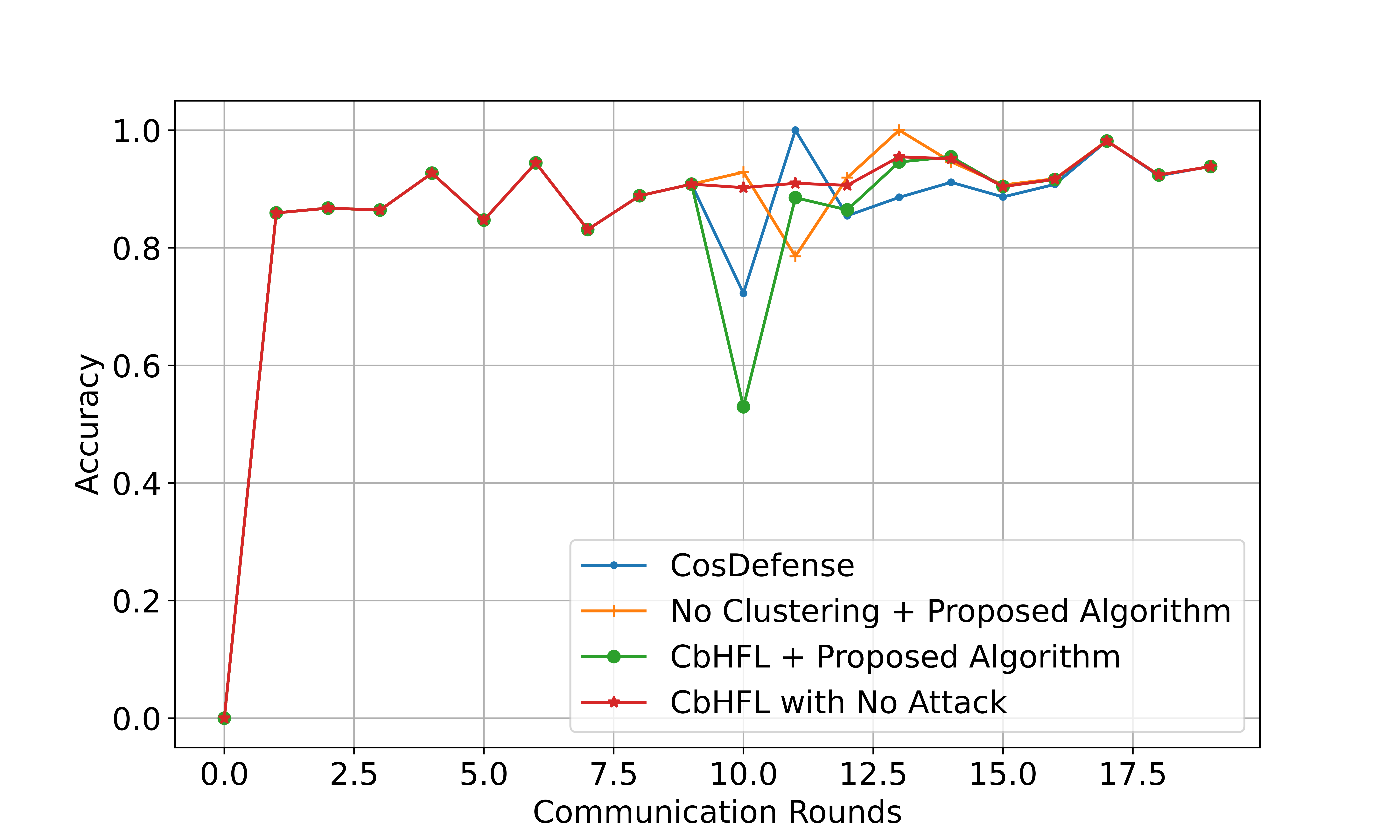}
         \caption{Mean= 0, Variance= 0.3}
         \label{fig: 10th-0-0.3}
     \end{subfigure}
     \vfill
     \begin{subfigure}{0.45\textwidth}
         \centering
         \includegraphics[width=\textwidth]{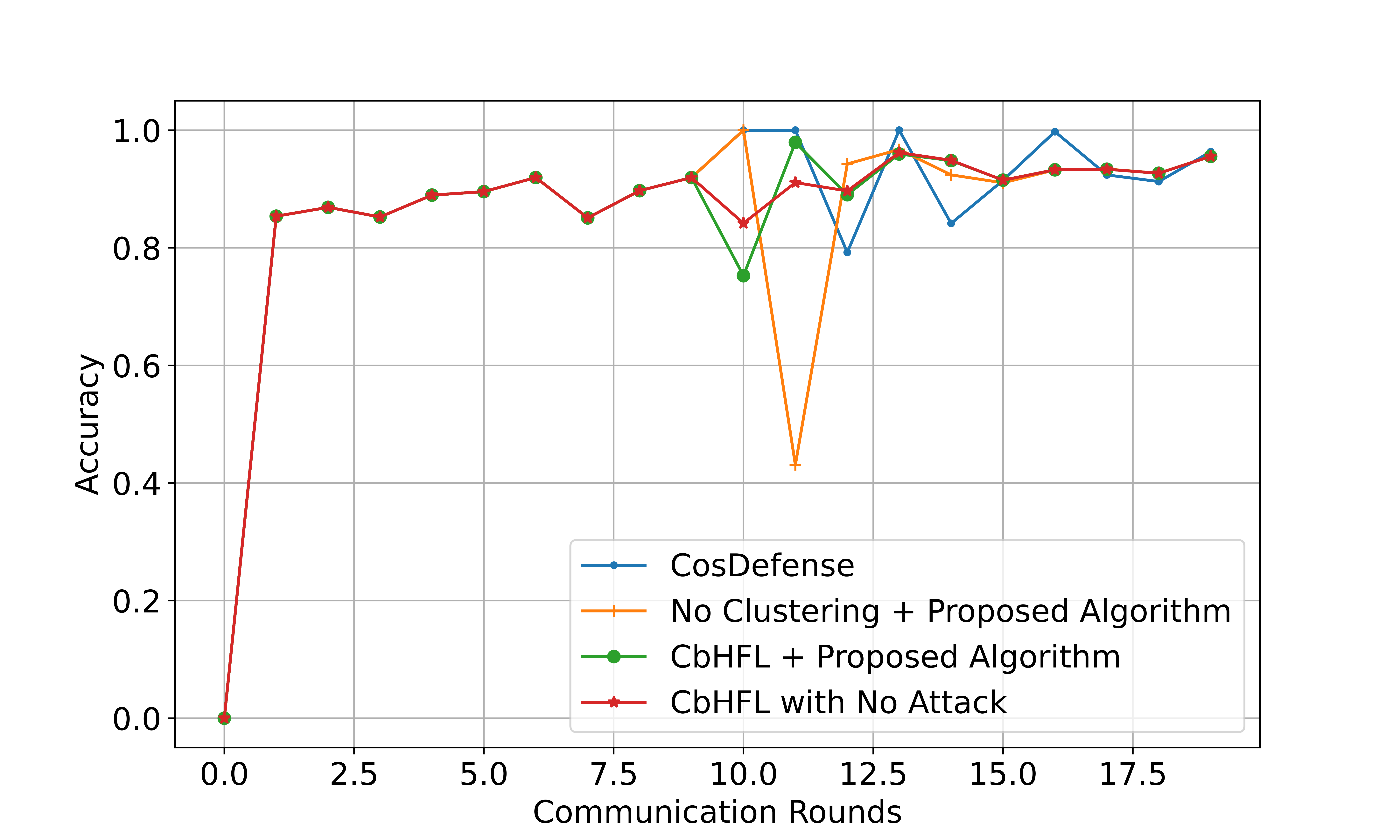}
         \caption{Mean= 1, Variance= 0.3}
         \label{fig: 10th-1-0.3}
     \end{subfigure}
     \vfill
     \begin{subfigure}{0.45\textwidth}
         \centering
         \includegraphics[width=\textwidth]{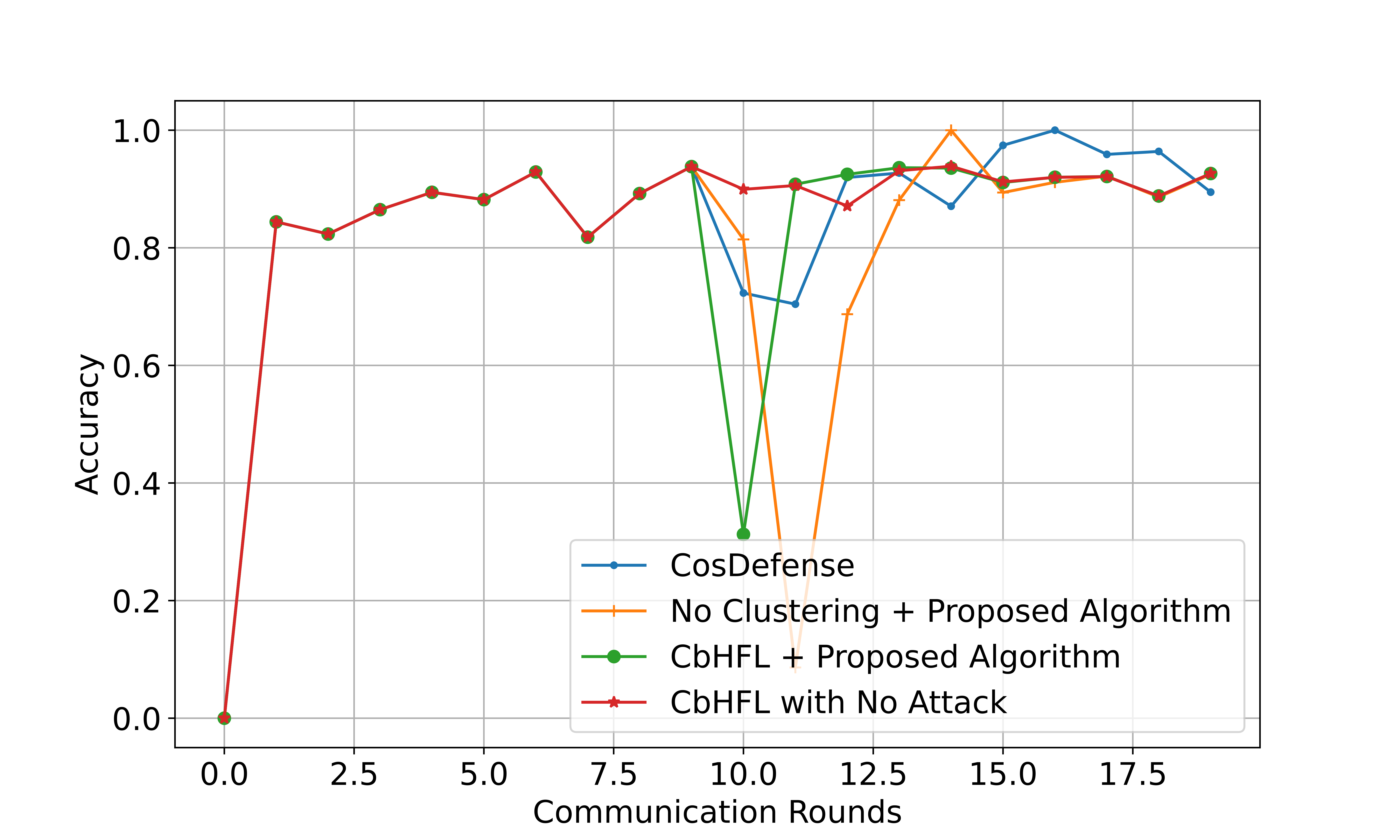}
         \caption{Mean= 2, Variance= 0.3}
         \label{fig: 10th-2-0.3}
     \end{subfigure}
        \caption{The accuracy in the EPC for various means and a variance of $0.3$ of the fake noise attack on the $10$th round.}
        \label{fig: EPC Accuracy for 10th Attack}
\end{figure}

\begin{figure}[!t]
     \centering
     \begin{subfigure}{0.45\textwidth}
         \centering
         \includegraphics[width=\textwidth]{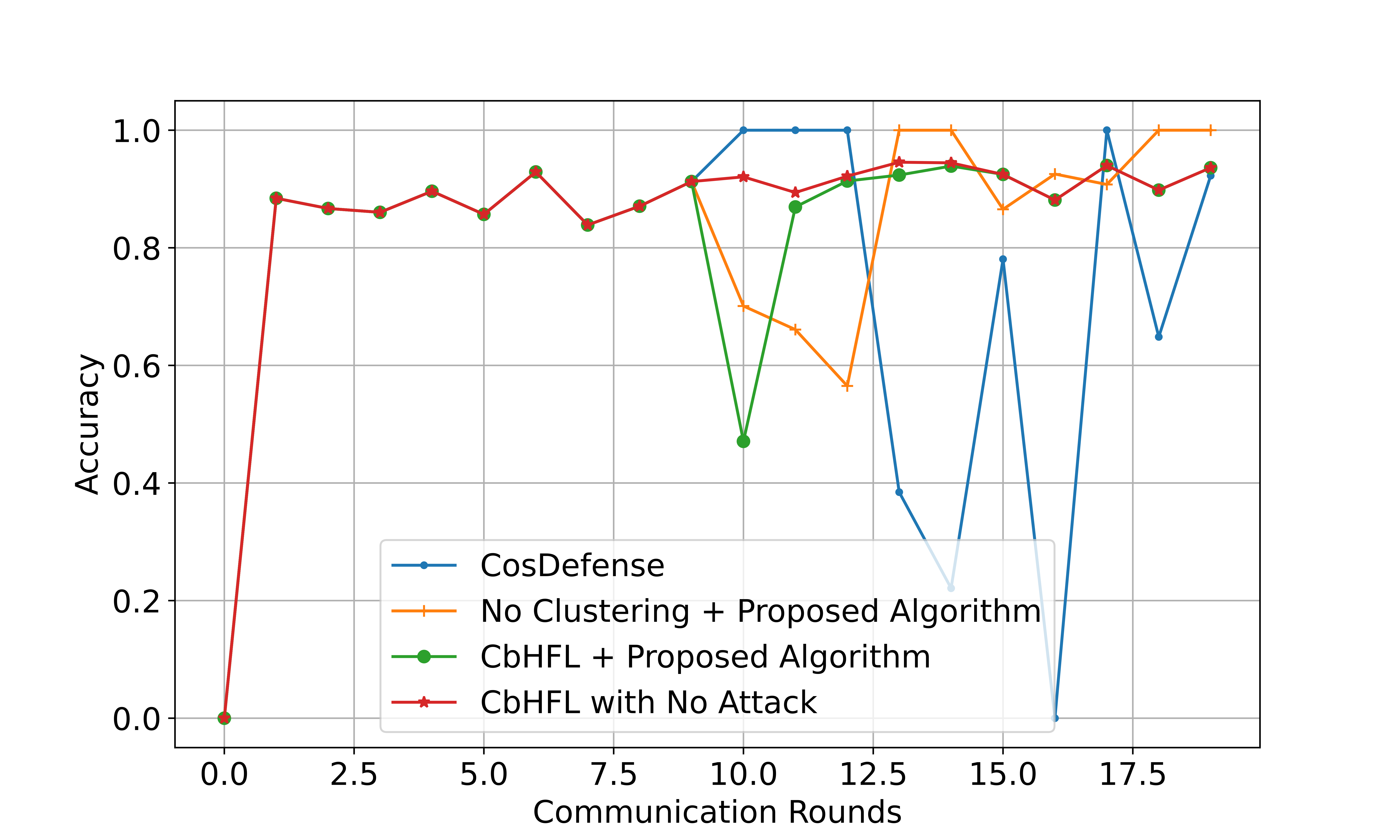}
         \caption{Mean= 0, Variance= 0.3}
         \label{fig: Count-0-0.3}
     \end{subfigure}
     \vfill
     \begin{subfigure}{0.45\textwidth}
         \centering
         \includegraphics[width=\textwidth]{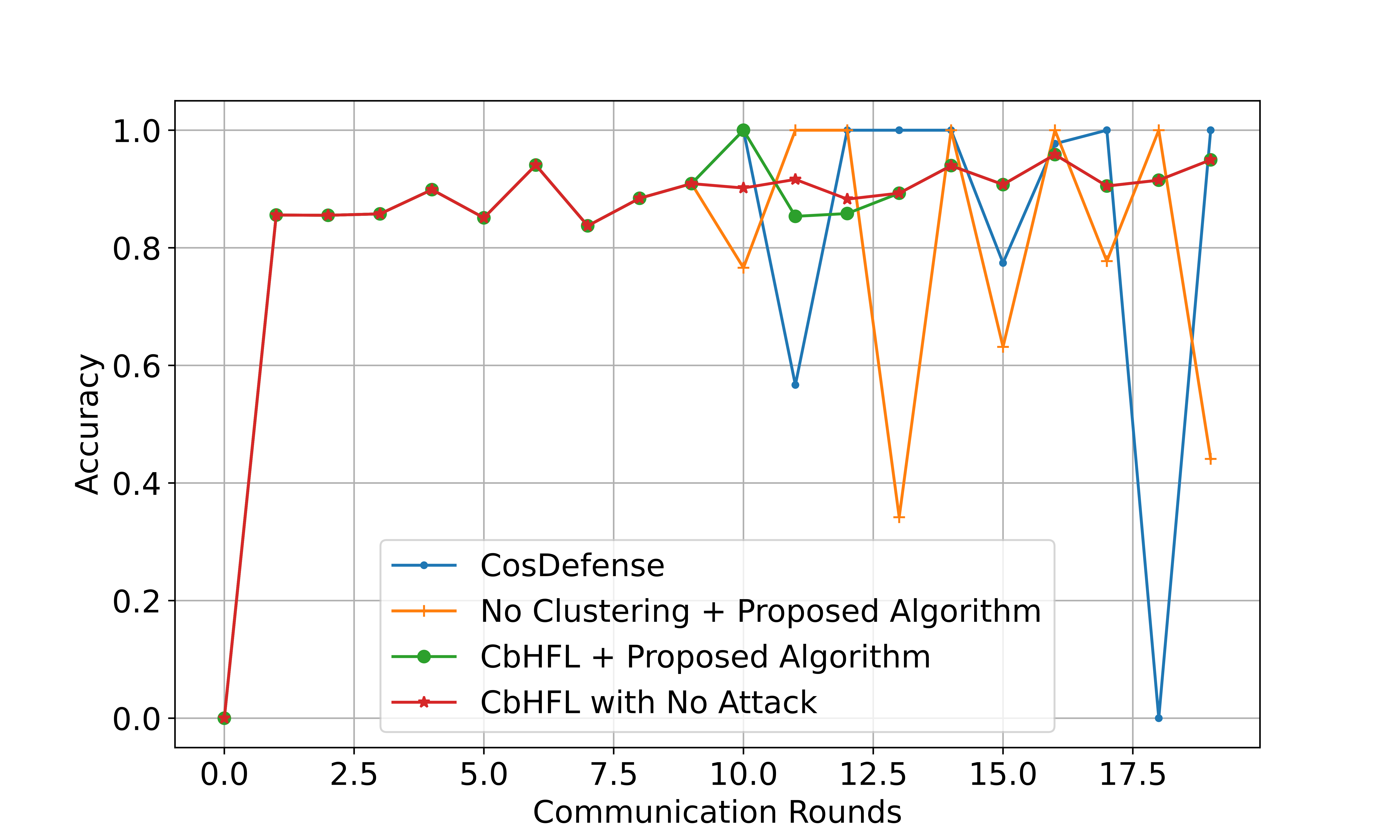}
         \caption{Mean= 1, Variance= 0.3}
         \label{fig: Count-1-0.3}
     \end{subfigure}
     \vfill
     \begin{subfigure}{0.45\textwidth}
         \centering
         \includegraphics[width=\textwidth]{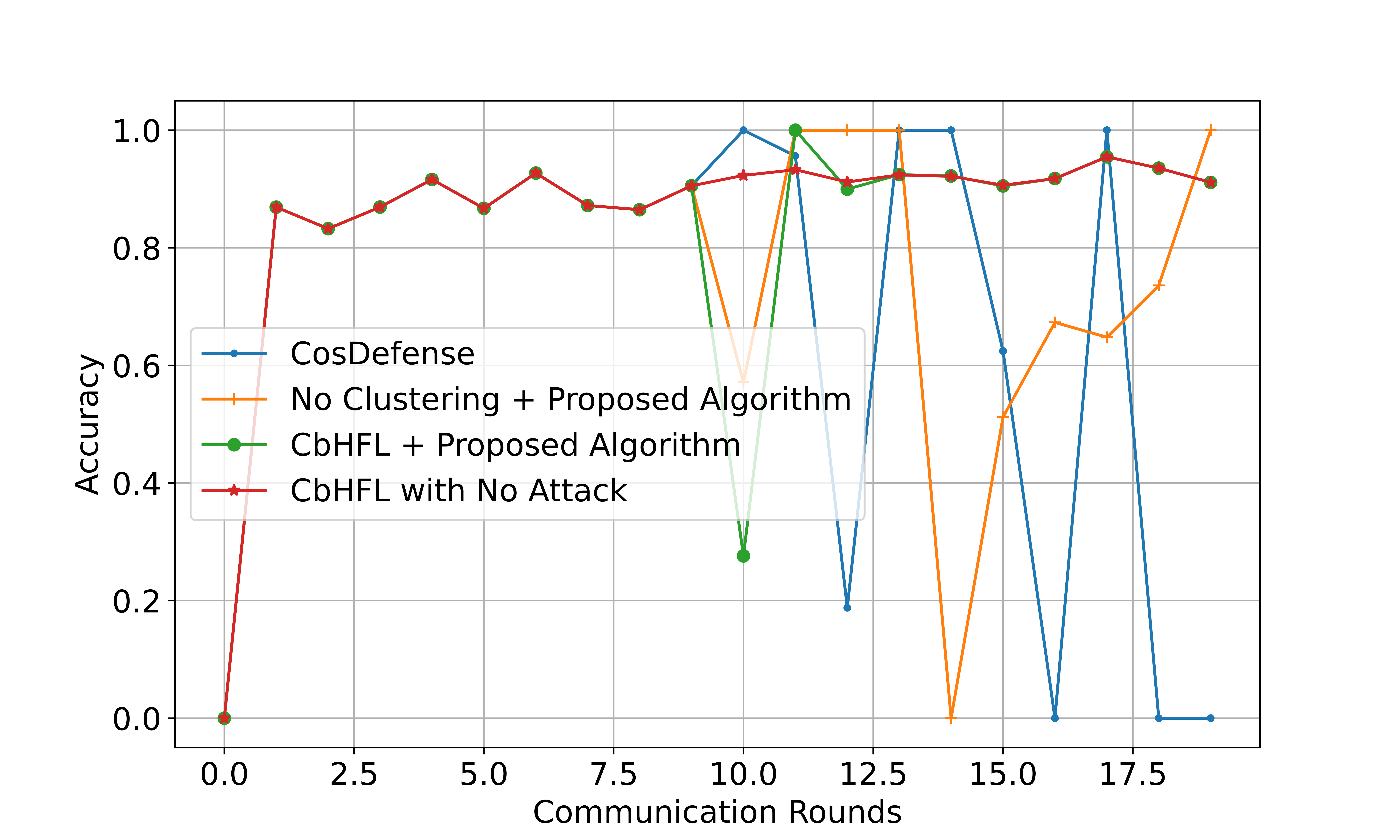}
         \caption{Mean= 2, Variance= 0.3}
         \label{fig: Count-2-0.3}
     \end{subfigure}
        \caption{The accuracy in the EPC for various means and a variance of $0.3$ of the fake noise attack continuously.}
        \label{fig: EPC Accuracy for Count Attack}
\end{figure}

Tables I and II show the convergence times of the proposed algorithm under different means and variances of the fake noise introduced by attackers, along with distinct values of $\epsilon$s. The convergence time is defined as the communication round at which the difference between the accuracy for three successive rounds falls below $\epsilon$. As the mean and variance of the attack increase, even if the attack occurs in the 10th round, CosDefense fails to converge the FL system, and the accuracy difference never falls below 0.1. Moreover, for continuous attacks, it is observed that neither CosDefense nor the no clustering scenario can achieve convergence. In contrast, the integration of CbHFL with the proposed algorithm demonstrates the ability to bring the system close to a state resembling the absence of any attack, highlighting the proposed algorithm's effectiveness.
The average convergence time for the CbHFL combined with the proposed algorithm is 33.1 for an attack occurring only in the 10th round and 38.5 for continuous attacks. This is in comparison to CbHFL without any attack, which has average convergence times of 26.3 and 24.4, respectively. Compared to the scenario with no attacks, the increased convergence time for the proposed algorithm can be attributed to the exclusion of vehicles exhibiting malicious behavior from the FL environment. This necessitates that the remaining vehicles be responsible for sustaining the FL process. Consequently, the learning burden is distributed among these remaining vehicles until new malicious users are identified and similarly addressed.

Figs. 1 and 2 illustrate the accuracy at the EPC under various means and a variance of $0.3$ for fake noise attacks occurring in the 10th round and continuously across different scenarios. Attacks initiated in the 10th round allow the CbHFL to stabilize the system rapidly when combined with the proposed algorithm. Even in scenarios where the attack is continuous, the proposed algorithm effectively stabilizes the FL by excluding vehicles attempting to disrupt the system with fake noise. In contrast, other methods fail to converge the FL under attacks characterized by high mean and variance, rendering the FL unsuccessful. Compared to scenarios without attacks, the observed fluctuations in the proposed algorithm's performance are due to the temporary inclusion of vehicles with malicious intent. These fluctuations persist until the system achieves convergence, as highlighted by the convergence times previously mentioned. Additionally, when comparing the system without clustering and CbHFL using the proposed algorithm, it is evident that the CbHFL structure inherently enhances system stability. Conversely, systems without clustering are more susceptible to destabilization by attacks.

\section{Conclusion}
The paper introduces a novel framework to enhance the security and integrity of HFL within VANETs by integrating dynamic client selection and anomaly detection. By prioritizing vehicles based on a reliability score derived from historical accuracy, contribution frequency, and anomaly records, the framework ensures the participation of only the most trustworthy vehicles in the FL process. Leveraging Hierarchical Federated Learning atop a Cluster-Based VANET approach, the algorithm integrates average relative speed and cosine similarity metrics for effective clustering and accelerated convergence within HFL contexts. Utilizing cosine similarity metrics for anomaly detection and tracking the history of malicious activities, the algorithm effectively isolates and neutralizes threats, thereby improving system stability and learning accuracy across various attack scenarios. Through comprehensive simulations, our framework, coupled with CbHFL, has been proven to enhance system stability and learning accuracy, even under various fake noise attack scenarios. Remarkably resilient, the algorithm excels in situations of both immediate and sustained attacks, surpassing existing solutions by adeptly sidelining malevolent vehicles to safeguard the learning process. 
Looking ahead, our research will aim to refine this algorithm to address a wider spectrum of attacks and assess its viability across various domains where FL plays a pivotal role.
\bibliographystyle{ieeetr}
\bibliography{references}

\end{document}